\documentclass[10pt,twocolumn,letterpaper]{article}

\usepackage{cvpr}
\usepackage{times}
\usepackage{epsfig}
\usepackage{graphicx}
\usepackage{amsmath}
\usepackage{amssymb}
\usepackage{color}
\usepackage{multicol, blindtext}
\usepackage{booktabs}
\usepackage{soul}
\usepackage{csquotes}
\usepackage[dvipsnames]{xcolor}
\definecolor{mypurple}{RGB}{160, 48, 160}
\definecolor{mygreen}{RGB}{84, 130, 53}
\definecolor{myorange}{RGB}{255, 210, 0}
\definecolor{myblue}{RGB}{0, 176, 240}
\definecolor{mybrown}{RGB}{190, 62, 19}

\usepackage[breaklinks=true,bookmarks=false]{hyperref}
\cvprfinalcopy 
\def\cvprPaperID{****} 

\begin{document}
\title{Single Underwater Image Restoration by Contrastive Learning}
\author{Junlin Han$^{1,2}$, Mehrdad Shoeiby$^{1}$, Tim Malthus$^{1}$, Elizabeth Botha$^{1}$,\\
Janet Anstee$^{1}$, Saeed Anwar$^{1,2}$, Ran Wei$^{1}$, Lars Petersson$^{1}$, Mohammad Ali Armin$^{1}$\\
$^{1}$DATA61-CSIRO, $^{2}$Australian National University
}
\maketitle
\begin{abstract}
Underwater image restoration attracts significant attention due to its importance in unveiling the underwater world. This paper elaborates on a novel method that achieves state-of-the-art results for underwater image restoration based on the unsupervised image-to-image translation framework. We design our method by leveraging from contrastive learning and generative adversarial networks to maximize mutual information between raw and restored images. Additionally, we release a large-scale real underwater image dataset to support both paired and unpaired training modules. Extensive experiments with comparisons to recent approaches further demonstrate the superiority of our proposed method.
\end{abstract}
\section{Introduction}
\label{sec:intro}
For marine science and ocean engineering, significant applications such as the surveillance of coral reefs, underwater robotic inspection, and inspection of submarine cables, require clear underwater images. However, raw underwater images with low visual quality can not meet the expectations. The quality of underwater images plays an essential role in scientific missions; thus, fast, accurate, and effective image restoration techniques need to be developed to improve the visibility, contrast, and color properties of underwater images for satisfactory visual quality.

In the underwater scene, visual quality is greatly affected by light refraction, absorption, and scattering. For instance, underwater images usually have a green-bluish tone since red light with longer wavelengths attenuates faster. Underwater image restoration is an ill-posed problem, which requires several parameters (\eg{} global background light and medium transmission map) that are mostly unavailable in practice. These parameters can be roughly estimated by employing priors and supplementary information. However, due to the diversity of water types and lighting conditions, conventional underwater image restoration methods fail to rectify the color of underwater images. 

Recent advances in deep learning demonstrate dramatic success in different fields. Learning-based models require a large-scale dataset for training, which is often difficult to obtain. Thus, most learning-based models use either small-scale real underwater images~\cite{li2019underwater,islam2020fast}, synthesized images~\cite{li2020underwater,fabbri2018enhancing}, or natural in-air images~\cite{li2018emerging} as either the source domain or target domain of the training set, instead of using the restored underwater images as the target domain. The aforementioned datasets are limited to capture natural variability in a wide range of water types.

To overcome the earlier discussed challenges, we construct a large-scale real underwater image dataset with accurate restored underwater images. We formulate the restoration problem as an image-to-image translation problem and propose a novel \textbf{C}ontrastive Under\textbf{W}ater \textbf{R}estoration approach (CWR). Given an underwater image as the input, CWR directly outputs a restored image showing the real color of underwater objects as if the image was taken in-air without any structure and content loss.

The main contribution of our work is summarized as:
\begin{itemize}
    \item We propose CWR, which leverages contrastive learning to maximize the mutual information between corresponding patches of the raw image and the restored image to capture the content and color feature correspondences between two image domains.
    \item We construct a large-scale, high-resolution underwater image dataset with real underwater images and restored images. This dataset supports both paired or unpaired training. Our code and dataset are available on \href{https://github.com/JunlinHan/CWR}{\textcolor{red}{GitHub}}.
\end{itemize}

\section{A Novel Dataset}
\label{sec:dataset}

Heron Island Coral Reef Dataset (HICRD) contains raw underwater images from eight sites with detailed metadata for each site, including water types, maximum dive depth, wavelength-dependent attenuation within the water column, and the camera model. According to raw images' depth information and the distance between objects and the camera, images with roughly the same depth and constant distance are labeled as good-quality. Images with sharp depth changes or distance changes are labeled as low-quality. We apply our imaging model described in section~\ref{sec:imaging} to good-quality images, producing corresponding restored images, and manually remove some restored images with non-satisfactory quality.

HICRD contains 6003 low-quality images, 3673 good-quality images, and 2000 restored images. We use low-quality images and restored images as the unpaired training set. In contrast, the paired training set contains good-quality images and corresponding restored images. The test set contains 300 good-quality images as well as 300 paired restored images as reference images. All images are in 1842 x 980 resolution.

\section{Underwater Imaging Model}
\label{sec:imaging}
Unlike the dehazing model~\cite{he2010single}, absorption plays a critical role in an underwater scenario. Each channel's absorption coefficient is wavelength-dependent, being the highest for red and the lowest for blue. A simplified underwater imaging model~\cite{serikawa2014underwater} can be formulated as:
\begin{equation}
    I^{c}(x) = J^{c}(x)t^{c}(x) + A^{c}(1-t^{c}(x)),~~~~c\in \{{r,g,b}\},
    \label{eq:haze}
\end{equation}
where $I(x)$ is the observed intensity, $J(x)$ is the scene radiance, and $A$ is the global atmospheric light. $t^{c}(x) = e^{-\beta^{c} d(x)}$ is the medium transmission describing $A(x)$ the portion of light that is not scattered, $\beta^{c}$ is the light attenuation coefficient and $d(x)$ is the distance between camera and object. Channels are in RGB space. 

\begin{figure}[!htb]
     \centering
     \includegraphics[scale=0.5]
     {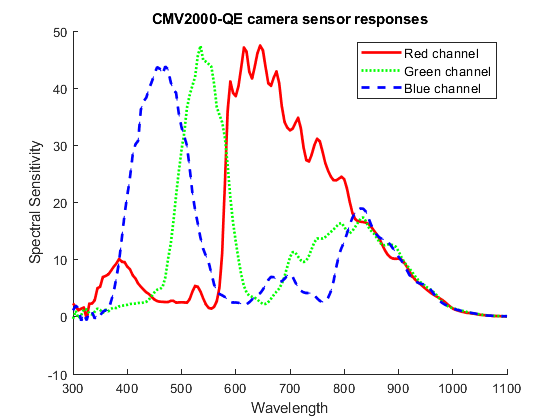}
     \caption{Camera sensor response for camera sensor type CMV2000-QE which is used in collecting real underwater images.}
     \label{fig:response}
\end{figure}

Transmittance is highly related to $\beta^{c}$, which is the light attenuation coefficient of each channel, and it is wavelength-dependent. Unlike previous work~\cite{peng2017underwater,chiang2011underwater}, instead of assigning a fixed wavelength for each channel containing bias (\eg, 600nm, 525nm, and 475nm for red, green, and blue), we employ the camera sensor response to conduct a more accurate estimation. Figure~\ref{fig:response} shows the camera sensor response of sensor type CMV2000-QE used in collecting the dataset.

The new total attenuation coefficient is estimated by:
\begin{equation}
    p^{c}=\int_{a}^{b}\beta^{\lambda}S^{c}(\lambda)d\lambda,
    \label{eq:response}
\end{equation}
where $p^{c}$ is the total attenuation coefﬁcient, $\beta^{\lambda}$ is the attenuation coefficient of each wavelength, and $S^{c}(\lambda)$ is the camera sensor response of each wavelength. Following the human visible spectrum, we set a = 400nm and b = 750nm to calculate the medium transmission for each channel. We modify $t^{c}(x)$ in equation~\ref{eq:haze} leading to a more accurate estimation: $t^{c}(x) = e^{-p^{c} d(x)}.$

It is challenging to measure the scene's actual distance from an individual image without a depth map. Instead of using a flawed estimation approach, we assume the distance between the scene and the camera to be small (1m - 5m) and manually assign a distance for each good-quality image.

The global atmospheric light, $A^{c}$ is usually assumed to be the pixel's intensity with the highest brightness value in each channel. However, this assumption often fails due to the presence of artificial lighting and self-luminous aquatic creatures. Since we have access to the diving depth, we can define $A^{c}$ as follows:
\begin{equation}
    A^{c} = e^{-p^{c \phi}},
    \label{eq:airlight}
\end{equation}
where $p^{c}$ is the total attenuation coefﬁcient, $\phi$ is the diving depth.

With the medium transmission and global atmospheric light, we can recover the scene radiance. The final scene radiance $J(x)$ is estimated as:
\begin{equation}
    J^{c}(x) = \frac{I_{c}(x)-A_{c}}{max(t_{c}(x),t_{0})}+A_{c}.
    \label{radiance}
\end{equation}
Typically, we choose $t_{0}$ = 0.1 as a lower bound. In practice, due to image the formulation's complexity, our imaging model may encounter information loss, \ie, the pixel intensity values of $J^{c}(x)$ are larger than 255 or less than 0. This problem is avoided by only mapping a selected range (13 to 255) of pixel intensity values from $I$ to $J$. However, outliers may still occur; we re-scale the whole pixel intensity values to enhance contrast and keep information lossless after restoration.
 \begin{table*}[!htbp]
  \centering
    \begin{tabular}{lcl}
    \toprule
    \textbf{Loss type}&\textbf{Equation No.}&\multicolumn{1}{c}{\textbf{Equation}}\cr
    \midrule
    Adversarial &6& 
    $\mathcal{L}_{\mathrm{GAN}}\left(G, D, X, Y\right) =\mathbb{E}_{y\sim{Y}}\left[(D(y))^2\right] +\mathbb{E}_{x\sim{X}}\left[\left(1-D(G(x))\right)^2]\right.$
    \cr
    Cross-entropy &7& 
    $ \ell\left(\boldsymbol{v}, \boldsymbol{v}^{+}, \boldsymbol{v}^{-}\right)= -\log( \frac{\exp \left(\boldsymbol cos({v},\boldsymbol{v}^{+}) / \tau\right)}{\exp \left(\boldsymbol cos({v},\boldsymbol{v}^{+}) / \tau\right)+\sum_{n=1}^{N} \exp \left(\boldsymbol cos({v},\boldsymbol{v}_{n}^{-}) / \tau\right)})$
    \cr   
    PatchNCE &8& 
    $\mathcal{L}_{\mathrm{PatchNCE}}(G,H,X)=
    \mathbb{E}_{\boldsymbol{x} \sim X} \sum_{l=1}^{L} \sum_{s=1}^{S_{l}} \ell\left(\hat{z}_{l}^{s}, \boldsymbol{z}_{l}^{s}, \boldsymbol{z}_{l}^{S \backslash s}\right)$
    \cr
    Identity &9& 
    $\mathcal{L}_{\text {Identity}}(G)=\mathbb{E}_{y \sim {Y} }\left[\|G(y)-y\|_{1}\right]$
    \cr   
    \bottomrule
    \end{tabular}
     \caption{Components of the full objective. We use least-square adversarial loss for equation 6. For equation 7, we use a noisy contrastive estimation framework to maximize the mutual information between inputs and outputs. The idea behind equation 7 is to correlate two signals, \ie, the \enquote{query} and its \enquote{positive} example, in contrast to other examples in the dataset (referred to as  \enquote{negatives}). We map query, positive, and $N$ negatives to $K$-dimensional vectors and denote them $v, v^{+} \in R^{K}$ and $ v^{-} \in R^{N \times K}$, respectively. Note that $v_{n}^{-} \in R^{K}$ denotes the n-th negative. We set up an $(N+1)$-way classification problem and compute the probability that a  \enquote{positive} is selected over \enquote{negatives}. This can be expressed as a cross-entropy loss where $\operatorname{cos}(\boldsymbol{u}, \boldsymbol{v})=\boldsymbol{u}^{\top} \boldsymbol{v} /\|\boldsymbol{u}\|\|\boldsymbol{v}\|$ denotes the cosine similarity between $\boldsymbol{u}$ and $\boldsymbol{v}$. $\tau$ denotes a temperature parameter to scale the distance between the query and other examples, we use 0.07 as default. For equation 8, We select $L$ layers from $G_{enc}(X)$ and send them to a projection head $H_{X}$, embedding one image to a stack of feature $\left\{\boldsymbol{z}_{l}\right\}_{L}=\left\{H^{l}\left(G_{\mathrm{enc}}^{l}(\boldsymbol{x})\right)\right\}_{L}$, where $G_{\mathrm{enc}}^{l}$ represents the output of $l$-th selected layers. After having a stack of features, each feature actually represents one patch from the image. We denote the spatial locations in each selected layer as $s \in \{1,...,S_{l}\}$, where $S_{l}$ is the number of spatial locations in each layer. We select a query each time, refer the corresponding feature \enquote{positive} as $\boldsymbol{z}_{l}^{s} \in \mathbb{R}^{C_{l}}$ and all other features (\enquote{negatives}) as $\boldsymbol{z}_{l}^{S \backslash s} \in \mathbb{R}^{\left(S_{l}-1\right) \times C_{l}}$, where $C_{l}$ is the number of channels in each layer. An Identity loss is introduced to prevent the generator from unnecessary changes. Equation 9 is a $\ell_1$ Identity loss preserving the fidelity.}
  \label{tab:equations}
\end{table*}

\section{Method}
Given two domains $\mathcal{X} \subset \mathbb{R}^{H \times W \times 3}$ and $\mathcal{Y} \subset \mathbb{R}^{H \times W \times 3}$ and a dataset of unpaired instances $X$ containing raw underwater images $x$ and $Y$ containing restored images $y$. We denote it  $X= \left\{x \in \mathcal{X} \right\}$ and $Y= \left\{y \in \mathcal{Y} \right\}$. We aim to learn a mapping $G : X\rightarrow Y$ to enable underwater image restoration.

\textbf{C}ontrastive Under\textbf{W}ater \textbf{R}estoration (CWR) has a generator $G$ as well as a discriminator $D$. $G$ enables the restoration process, and $D$ ensures that the images generated by $G$ are undistinguished to domain $Y$ in principle. The first half of the generator is defined as an encoder while the second half is a decoder, presented as $G_{enc}$ and $G_{dec}$ respectively.

We extract features from several layers of the encoder and forward them to a two-layer MLP projection head $H$. Such a projection head learns to project the extracted features from the encoder to a stack of features. CWR combines three losses, including Adversarial loss, PatchNCE loss, and Identity loss. The details of our objective are described below.

The restored image should be realistic ($\mathcal{L}_{GAN}$), and patches in the corresponding raw and restored images should share some correspondence ($\mathcal{L}_{\mathrm{PatchNCE}}$). The restored image should have an identical structure to the raw image. In contrast, the colors are the true colors of scenes 
($\mathcal{L}_{\mathrm{Identity}}$). The full objective is:
\begin{equation}
\begin{aligned}
\mathcal{L}(G,D,H)
&=\lambda_{GAN}\mathcal{L}_{GAN}(G,D,X)\\ 
&+\lambda_{NCE}\mathcal{L}_{\mathrm{PatchNCE}}(G, H, X)\\
&+\lambda_{IDT}\mathcal{L}_{\text {Identity}}(G).
\end{aligned}
\end{equation}
We set $\lambda_{GAN}$ = 1, $\lambda_{NCE}$ = 1, and $\lambda_{IDT}$ = 10. The details of each component are elaborated in Table~\ref{tab:equations}.
\section{Experiments}
\subsection{Baselines and Training Details}
We compare CWR to several state-of-the-art baselines from different views, including image-to-image translation approaches (CUT~\cite{park2020contrastive} and CycleGAN~\cite{CycleGAN2017}), underwater image enhancement methods (UWCNN~\cite{li2020underwater}, Retinex~\cite{fu2014retinex} and Fusion~\cite{ancuti2017color}), and underwater image restoration methods (DCP~\cite{he2010single}, IBLA~\cite{peng2017underwater}). We use the pre-trained UWCNN model with water type-3, which is close to our dataset.

We train CWR, CUT, and CycleGAN for 100 epochs with the same learning rate of 0.0002. The learning rate decays linearly after half epochs. We load all images in 800x800 resolution, and randomly crop them into 512x512 patches during training. We load test images in 1680x892 resolution for all methods. The architecture of CWR is inspired by CUT, a Resnet-based generator with nine residual blocks and a PatchGAN discriminator. We employ spectral normalization for discriminator and instance normalization for generator. The batch size is 1 and the optimizer is Adam.
\begin{figure*}[!htbp]
  \begin{minipage}[t]{0.09\linewidth} 
    \centering 
    \text{\scriptsize Input}
  \end{minipage} 
    \begin{minipage}[t]{0.09\linewidth} 
    \centering 
    \text{\scriptsize CUT \cite{park2020contrastive}}
    \end{minipage}
    \begin{minipage}[t]{0.09\linewidth} 
    \centering 
    \text{\scriptsize CycleGAN \cite{CycleGAN2017}}
    \end{minipage}
    \begin{minipage}[t]{0.09\linewidth} 
    \centering 
    \text{\scriptsize UWCNN \cite{li2020underwater}}
    \end{minipage}
    \begin{minipage}[t]{0.09\linewidth} 
    \centering 
    \text{\scriptsize Retinex \cite{fu2014retinex}}
    \end{minipage}
    \begin{minipage}[t]{0.09\linewidth} 
    \centering 
    \text{\scriptsize Fusion \cite{ancuti2017color}}
    \end{minipage}
    \begin{minipage}[t]{0.09\linewidth} 
    \centering 
    \text{\scriptsize DCP \cite{he2010single}}
    \end{minipage}
    \begin{minipage}[t]{0.09\linewidth} 
    \centering 
    \text{\scriptsize IBLA \cite{peng2017underwater}}
    \end{minipage}
    \begin{minipage}[t]{0.09\linewidth} 
    \centering 
    \text{\scriptsize CWR (ours)}
    \end{minipage}
     \begin{minipage}[t]{0.09\linewidth} 
    \centering 
    \text{\scriptsize Ground Truth}
    \end{minipage}   
  \\
  \begin{minipage}[t]{0.09\linewidth} 
    \centering 
    \includegraphics[width=0.65in, height=0.37in]{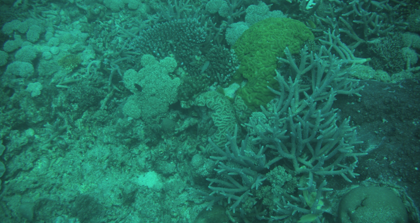}
  \end{minipage} 
    \begin{minipage}[t]{0.09\linewidth} 
    \centering 
    \includegraphics[width=0.65in, height=0.37in]{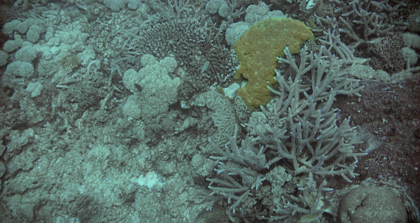}
  \end{minipage} 
     \begin{minipage}[t]{0.09\linewidth} 
    \centering 
    \includegraphics[width=0.65in, height=0.37in]{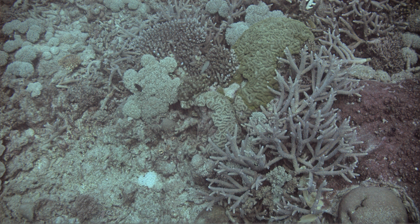}
  \end{minipage}  
     \begin{minipage}[t]{0.09\linewidth} 
    \centering 
    \includegraphics[width=0.65in, height=0.37in]{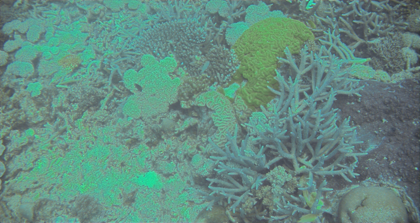}
  \end{minipage}  
     \begin{minipage}[t]{0.09\linewidth} 
    \centering 
    \includegraphics[width=0.65in, height=0.37in]{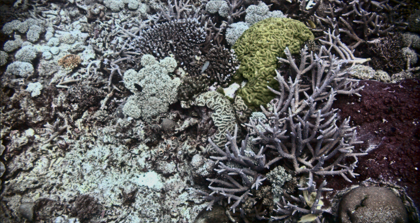}
  \end{minipage}  
     \begin{minipage}[t]{0.09\linewidth} 
    \centering 
    \includegraphics[width=0.65in, height=0.37in]{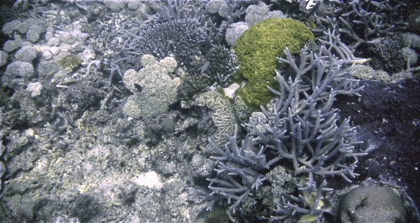}
  \end{minipage}  
     \begin{minipage}[t]{0.09\linewidth} 
    \centering 
    \includegraphics[width=0.65in, height=0.37in]{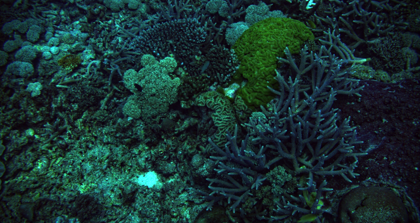}
  \end{minipage}  
     \begin{minipage}[t]{0.09\linewidth} 
    \centering 
    \includegraphics[width=0.65in, height=0.37in]{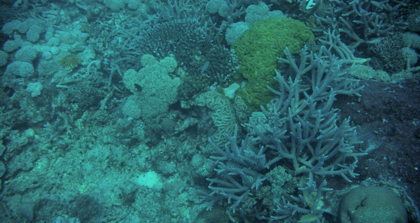}
  \end{minipage}  
     \begin{minipage}[t]{0.09\linewidth} 
    \centering 
    \includegraphics[width=0.65in, height=0.37in]{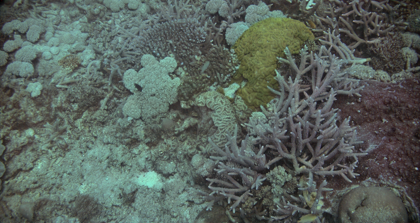}
  \end{minipage}  
     \begin{minipage}[t]{0.09\linewidth} 
    \centering 
    \includegraphics[width=0.65in, height=0.37in]{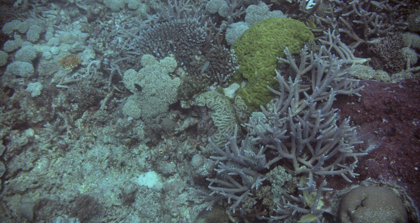}
  \end{minipage}  
  \\
   \begin{minipage}[t]{0.09\linewidth} 
    \centering 
    \includegraphics[width=0.65in, height=0.37in]{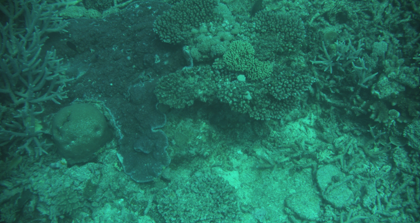}
  \end{minipage} 
    \begin{minipage}[t]{0.09\linewidth} 
    \centering 
    \includegraphics[width=0.65in, height=0.37in]{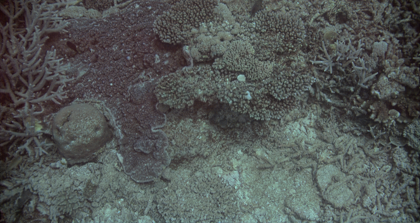}
  \end{minipage} 
     \begin{minipage}[t]{0.09\linewidth} 
    \centering 
    \includegraphics[width=0.65in, height=0.37in]{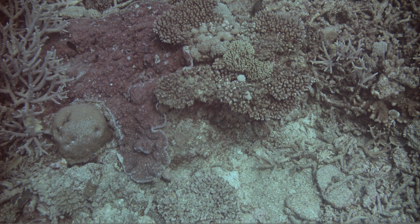}
  \end{minipage}  
     \begin{minipage}[t]{0.09\linewidth} 
    \centering 
    \includegraphics[width=0.65in, height=0.37in]{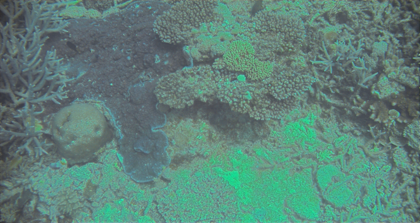}
  \end{minipage}  
     \begin{minipage}[t]{0.09\linewidth} 
    \centering 
    \includegraphics[width=0.65in, height=0.37in]{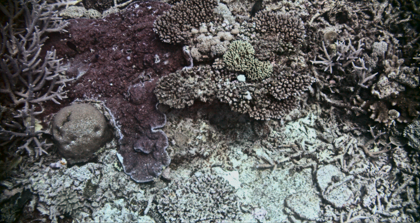}
  \end{minipage}  
     \begin{minipage}[t]{0.09\linewidth} 
    \centering 
    \includegraphics[width=0.65in, height=0.37in]{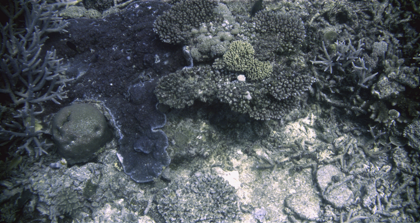}
  \end{minipage}  
     \begin{minipage}[t]{0.09\linewidth} 
    \centering 
    \includegraphics[width=0.65in, height=0.37in]{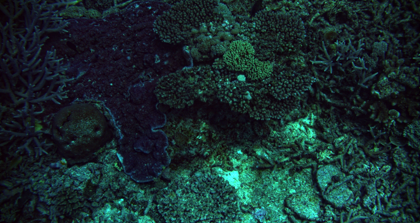}
  \end{minipage}  
     \begin{minipage}[t]{0.09\linewidth} 
    \centering 
    \includegraphics[width=0.65in, height=0.37in]{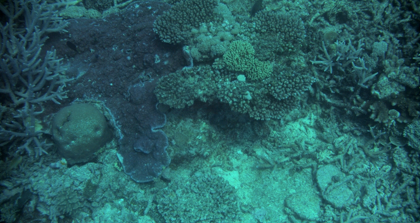}
  \end{minipage}  
     \begin{minipage}[t]{0.09\linewidth} 
    \centering 
    \includegraphics[width=0.65in, height=0.37in]{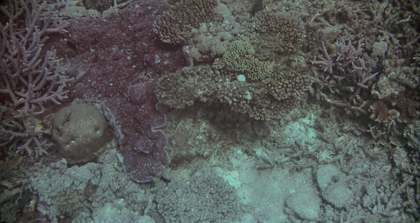}
  \end{minipage}  
     \begin{minipage}[t]{0.09\linewidth} 
    \centering 
    \includegraphics[width=0.65in, height=0.37in]{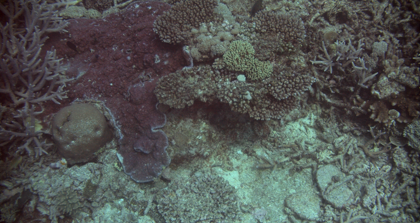}
  \end{minipage}  
  \\
     \begin{minipage}[t]{0.09\linewidth} 
    \centering 
    \includegraphics[width=0.65in, height=0.37in]{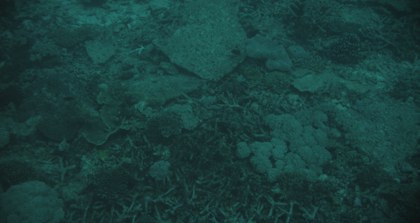}
  \end{minipage} 
    \begin{minipage}[t]{0.09\linewidth} 
    \centering 
    \includegraphics[width=0.65in, height=0.37in]{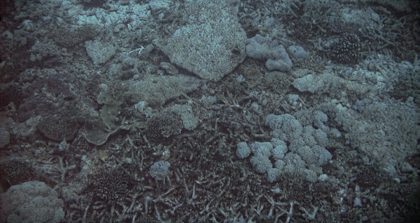}
  \end{minipage} 
     \begin{minipage}[t]{0.09\linewidth} 
    \centering 
    \includegraphics[width=0.65in, height=0.37in]{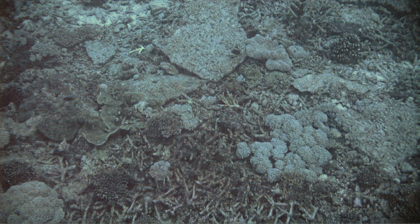}
  \end{minipage}  
     \begin{minipage}[t]{0.09\linewidth} 
    \centering 
    \includegraphics[width=0.65in, height=0.37in]{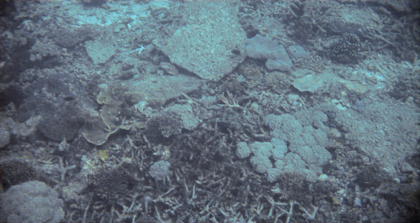}
  \end{minipage}  
     \begin{minipage}[t]{0.09\linewidth} 
    \centering 
    \includegraphics[width=0.65in, height=0.37in]{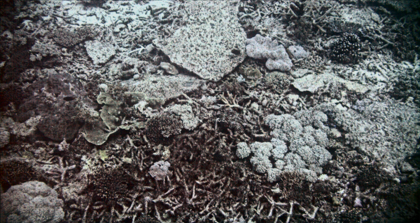}
  \end{minipage}  
     \begin{minipage}[t]{0.09\linewidth} 
    \centering 
    \includegraphics[width=0.65in, height=0.37in]{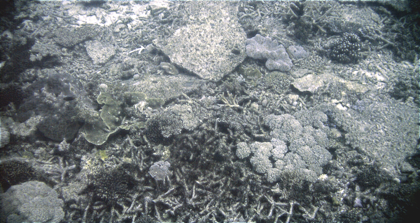}
  \end{minipage}  
     \begin{minipage}[t]{0.09\linewidth} 
    \centering 
    \includegraphics[width=0.65in, height=0.37in]{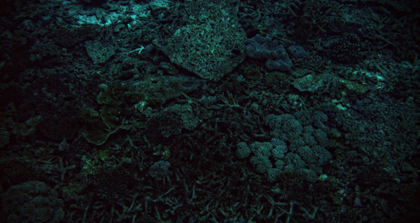}
  \end{minipage}  
     \begin{minipage}[t]{0.09\linewidth} 
    \centering 
    \includegraphics[width=0.65in, height=0.37in]{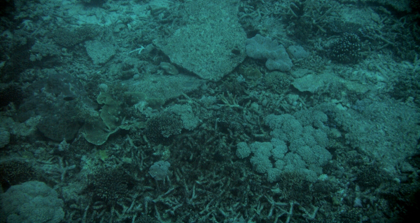}
  \end{minipage}  
     \begin{minipage}[t]{0.09\linewidth} 
    \centering 
    \includegraphics[width=0.65in, height=0.37in]{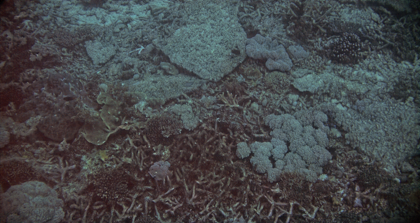}
  \end{minipage}  
     \begin{minipage}[t]{0.09\linewidth} 
    \centering 
    \includegraphics[width=0.65in, height=0.37in]{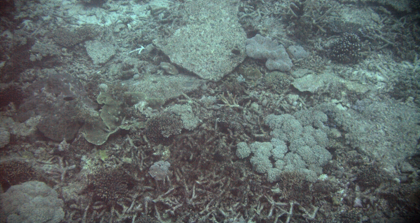}
  \end{minipage}  
  \\
       \begin{minipage}[t]{0.09\linewidth} 
    \centering 
    \includegraphics[width=0.65in, height=0.37in]{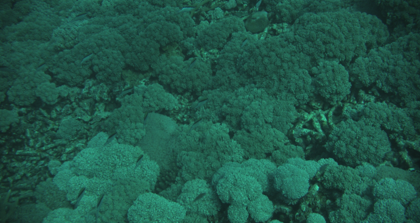}
  \end{minipage} 
    \begin{minipage}[t]{0.09\linewidth} 
    \centering 
    \includegraphics[width=0.65in, height=0.37in]{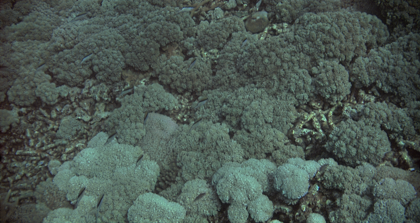}
  \end{minipage} 
     \begin{minipage}[t]{0.09\linewidth} 
    \centering 
    \includegraphics[width=0.65in, height=0.37in]{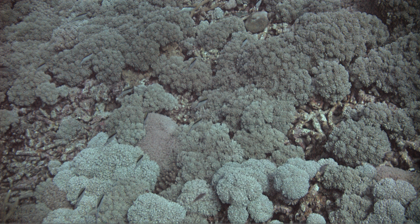}
  \end{minipage}  
     \begin{minipage}[t]{0.09\linewidth} 
    \centering 
    \includegraphics[width=0.65in, height=0.37in]{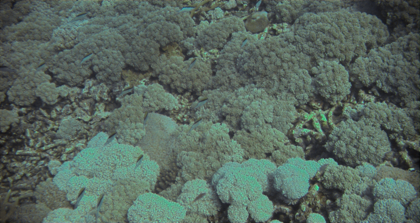}
  \end{minipage}  
     \begin{minipage}[t]{0.09\linewidth} 
    \centering 
    \includegraphics[width=0.65in, height=0.37in]{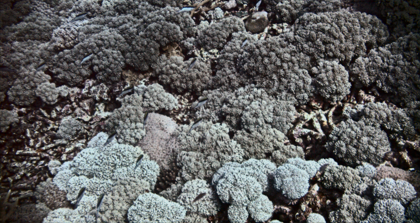}
  \end{minipage}  
     \begin{minipage}[t]{0.09\linewidth} 
    \centering 
    \includegraphics[width=0.65in, height=0.37in]{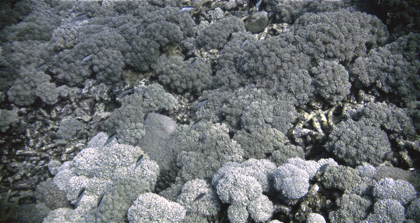}
  \end{minipage}  
     \begin{minipage}[t]{0.09\linewidth} 
    \centering 
    \includegraphics[width=0.65in, height=0.37in]{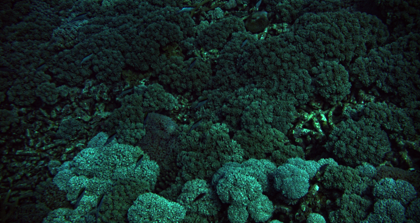}
  \end{minipage}  
     \begin{minipage}[t]{0.09\linewidth} 
    \centering 
    \includegraphics[width=0.65in, height=0.37in]{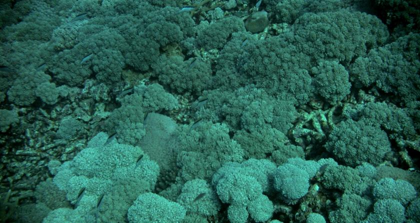}
  \end{minipage}  
     \begin{minipage}[t]{0.09\linewidth} 
    \centering 
    \includegraphics[width=0.65in, height=0.37in]{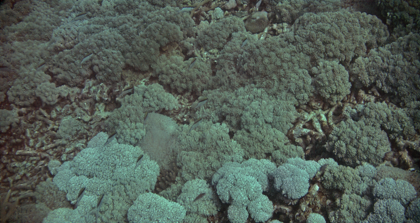}
  \end{minipage}  
     \begin{minipage}[t]{0.09\linewidth} 
    \centering 
    \includegraphics[width=0.65in, height=0.37in]{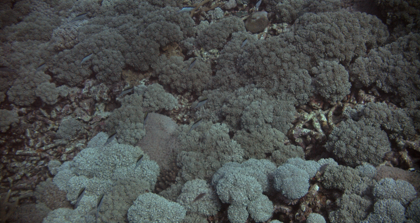}
  \end{minipage} 
  \caption{Qualitative results on the test set of HICRD. CWR shows visual satisfactory results without content and structure loss. }
  \label{fig:result1}
  \vspace{-5mm}
\end{figure*}
\subsection{Evaluation Protocol and Results}
To fully measure the performance of different methods, we employ three full-reference metrics: mean-square error (MSE), peak signal-to-noise ratio (PSNR), and structural similarity index (SSIM) as well as a non-reference metric designed for underwater images: Underwater Image Quality Measure (UIQM)~\cite{panetta2015human}. A higher UIQM score suggests the result is more consistent with human visual perception. We additionally use Fréchet Inception Distance (FID)~\cite{TTUR} to measure the quality of generated images. A lower FID score means generated images tend to be more realistic.

\begin{table}[!htbp]
  \centering
  \resizebox{\columnwidth}{!}{
    \begin{tabular}{lccccc}
    \toprule
    \cmidrule(lr){2-6}
    \textbf{Method} &\textbf{MSE$\downarrow$} &\textbf{PSNR$\uparrow$} &\textbf{SSIM$\uparrow$} &\textbf{UIQM$\uparrow$} &\textbf{FID$\downarrow$}   \cr
    \midrule
    CUT~\cite{park2020contrastive}  &\underline{170.27}&\underline{26.30}&\underline{0.796}&5.26&22.35\cr
    CycleGAN~\cite{CycleGAN2017}    &448.16    &21.81&0.591&5.27&\textbf{16.74}\cr
    \cmidrule(lr){1-6}     
    UWCNN~\cite{li2020underwater}   &775.81  &20.20&0.754&4.18&33.43\cr
    Retinex~\cite{fu2014retinex}    &1227.19  &17.36&0.722&\textbf{5.43}&71.90\cr  
    Fusion~\cite{ancuti2017color}   &1238.60  &17.53&0.783&\underline{5.33}&58.57\cr
    \cmidrule(lr){1-6} 
    DCP~\cite{he2010single}         &2548.20  &14.27&0.534&4.49&37.52\cr  
    IBLA~\cite{peng2017underwater}  &803.89 &19.42&0.459&3.63&23.06\cr  
    \cmidrule(lr){1-6}  
    CWR (ours) &\textbf{127.23}&\textbf{26.88}&\textbf{0.834}&5.25&\underline{18.20}\cr 
    \bottomrule
    \end{tabular}
    }
     \caption{Comparisons to baselines on HICRD dataset. We show five metrics for all methods. CWR is in the first place for MSE, PNSR, and SSIM while the second place for FID.}
     \label{tab:3compare}
\end{table}
Table~\ref{tab:3compare} provides quantitative evaluation, where no method always wins in terms of all metrics. However, CWR performs stronger than all the baselines. Figure~\ref{fig:result1} presents the \underline{randomly} selected qualitative results. Conventional methods produce blurry and unrealistic results, while learning-based methods tend to rectify the distorted color successfully. CWR performs better than other learning-based methods in keeping with the structure and content of the restored images identical to raw images with negligible artifacts.

\section{Conclusion}
This paper presents an underwater image dataset HICRD that offers large-scale underwater images and restored images to enable a comprehensive evaluation of existing underwater image enhancement \& restoration methods. We believe that HICRD will make a significant advancement for the use of learning-based methods. A novel method, CWR employing contrastive learning is proposed to capitalize on HICRD. Experimental results show that CWR significantly performs better than several conventional methods while showing more desirable results compared to learning-based methods.  
\small{
\bibliographystyle{ieee_fullname}
\bibliography{egbib}}

\end{document}